\pdfoutput=1

\documentclass[11pt]{article}

\usepackage[]{emnlp2021}

\usepackage{times}
\usepackage{latexsym}
\usepackage[T1]{fontenc}
\usepackage{amsthm}
\usepackage{amsmath}
\usepackage[ruled,linesnumbered]{algorithm2e}
\usepackage{booktabs}
\usepackage{graphicx}
\usepackage{tabularx}
\usepackage{flushend}
\newcolumntype{Y}{>{\centering\arraybackslash}X}
\newtheorem{definition}{Definition}[section]
\usepackage[utf8]{inputenc}
\usepackage{microtype}

\def\paperDraft{}
\usepackage[normalem]{ulem}
\ifdefined\paperDraft

\title{CAPE: Context-Aware Private Embeddings for Private Language Learning}

\author{Richard Plant \\
  Edinburgh Napier University \\
  \texttt{r.plant@napier.ac.uk} \\ \And
  Valerio Giuffrida\\
  Edinburgh Napier University\\
  \texttt{v.giuffrida@napier.ac.uk}  \\ \AND
  Dimitra Gkatzia \\
  Edinburgh Napier University \\
  \texttt{d.gkatzia@napier.ac.uk}}

\begin{document}
\maketitle
\begin{abstract}
Deep learning-based language models have achieved state-of-the-art results in a number of applications including sentiment analysis, topic labelling, intent classification and others. Obtaining text representations or embeddings using these models presents the possibility of encoding personally identifiable information learned from language and context cues that may present a risk to reputation or privacy. To ameliorate these issues, we propose \textit{Context-Aware Private Embeddings (CAPE)}, a novel approach which preserves privacy during training of embeddings. To maintain the privacy of text representations, CAPE applies calibrated noise through differential privacy, preserving the encoded semantic links while obscuring sensitive information. In addition, CAPE employs an adversarial training regime that obscures identified private variables. Experimental results demonstrate that the proposed approach reduces private information leakage better than either single intervention.
\end{abstract}

\section{Introduction}
Deep learning has provided remarkable advances in language understanding and modelling tasks in recent years \citep{Vaswani2017transformer, Devlin2019bert, Brown2020gpt3}. However, this increased utility may harm user privacy, as neural models trained with datasets containing personal identifiable information can unintentionally leak information that users may prefer to keep private \citep{Carlini2019secret, Song2017secret}. Even seemingly innocuous collections of metadata \cite{Xu2008deanon} such as data provided by the users (e.g. at registration time on social media) or data which has been cleansed of identifying attributes \cite{Sun2012deanon}, can provide \textit{latent} information for the re-identification of participants.

Using social media data can also raise ethical considerations \cite{Townsend:2016}. Users may have edited or deleted posts  that models continue to rely on in existing datasets, and may unintentionally reveal information they would rather keep private \citep{Bartunov2012, Pontes2012, Goga2013}. Research has shown practical attacks that exploit trained models to establish whether a particular individual formed part of a model's training dataset, in an attack known as membership inference \citep{Leino2020inference, Truex2018inference}. Personally identifiable attributes such as age, gender, or location can be reliably reconstructed given the output of such a model \citep{Fredrikson2015inference, Zhang2020inference}. Neural representations of input data, including language embeddings, have proven to be a vulnerability for these inferences \citep{Song2020inference}, thus privacy-preserving techniques should be applied to these text representations when they form part of a machine learning pipeline.

To minimise the risk of such attacks in uncovering sensitive information, previous work has employed an adversarial training objective \citep{Coavoux2018inference, Li2018adversarial} by modifying the loss function of the model to impose a penalty when a simulated attacker task, such as predicting a private variable from the input sequence, performs well. However, this approach provides no formal privacy guarantees nor privacy loss accounting system. \citet{Phan2019adversarial} proposed an approach which implements classical differential privacy in an adversarial learning paradigm, however, this work relies on adversarial objectives to promote robustness to adversarial samples rather than privacy.

Providing a privacy guarantee leads to the notion of differential privacy (DP), as defined by \citet{Dwork2013foundations}. This definition quantifies privacy loss as the maximum possible deviation between the same aggregate function applied to two datasets which differ only in a single record, which can be expressed by the variable $\epsilon$. 

\theoremstyle{definition} 
\begin{definition}[$\epsilon$-differential privacy]
The level of private information leaked by a computation $M$ can be expressed by the variable $\epsilon$ where for any two data sets $A$ and $B$, and any set of possible outputs $S \subset Range(M)$,
\begin{equation*}
\resizebox{0.99\hsize}{!}{$[M(A) \in S] \leq Pr[M(B) \in S] \times exp(\epsilon \times |A \oplus B|)$}
\end{equation*}
\label{def:epsilon_privacy}
\end{definition}

This notion of $\epsilon$-differential privacy has been extended to text embeddings through the application of calibrated noise \citep{Fernandes2019, Beigi2019embeds}. \citet{Lyu2020perturb} proposed a method based on local differential privacy---an extension to the schema under which noise is applied to the input data before it leaves the user's device and is encountered by the model owner---producing a private representation which can be sent to a server for classification. However, this approach uses simulated attacker performance as a test benchmark for private information leakage, rather than during training to improve privacy outcomes.

\paragraph{Contributions:} In this work, we propose an approach that combines perturbed pre-trained embeddings with a privacy-preserving adversarial training function that helps preserving the encoded semantic links in the input text while obscuring sensitive information. We demonstrate that our approach achieves comparable task performance against a competitive baseline while preserving privacy. We experiment with a dataset that contains personally identifiable information namely gender, location and birth year. To minimize harm, we experiment with a publicly available English-language dataset \citep{Hovy2015}. Specifically: 

\begin{itemize}
    \item We introduce CAPE, "Context-Aware Private Embeddings", an approach that applies both DP-compliant perturbations and an adversarial learning objective to privatize the embedding outputs of pre-trained language models.
    \item We establish metrics for testing the privacy result of our system against non-DP-compliant models by offering an empirical framework for determining the level of success of simulated attacks.
    \item We find that attacker inferences demonstrate differing levels of accuracy depending on the type of private attribute targeted.
    \item We establish superior privacy outcomes for our method compared to a sample adversarial learning approach \citep{coavoux_privacy-preserving_2018} and a perturbation-only method \citep{Lyu2020perturb} representing the dominant approaches currently applied to other task domains.
\end{itemize}

\section{Methodology}
\label{sec:method}

\begin{figure}[ht]
    \centering
    \includegraphics[width=\columnwidth]{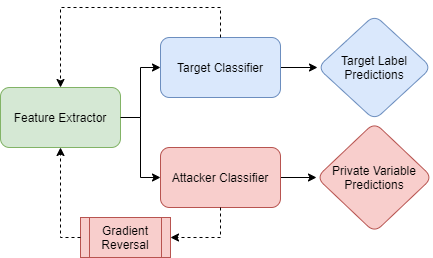}
    \caption{CAPE model diagram. Solid lines indicate data flow, dotted lines indicate gradient updates.}
    \label{fig:model}
\end{figure}

We consider the possibility that an attacker may have access to the intermediate feature representations extracted from text from a published language model along with a supervision signal that may allow them to train a model to recover private information about the text author, possibly garnered from access to a secondary data source as demonstrated in \citet{Narayanan2008deanon} and \citet{Carlini:2020}. To mitigate this risk, we introduce a DP-compliant layer to the feature extractor that perturbs the representations by adding calibrated noise. We train a second classifier to predict known private variables in addition to our main target task classifier, then pass the error gradient from the secondary classifier through a reversal layer to promote embedding invariance to the private features. Figure \ref{fig:model} shows the system architecture.

\subsection{Task}

We experiment with multi-class sentiment analysis on the UK section of the Trustpilot dataset \citep{Hovy2015}, which provides text reviews with an attached numerical rating from 1-5 as well as three demographic attributes: gender, location and birth year. Sentiment analysis from text reviews represents a popular task to which pre-trained language models are well suited. We use the gender as reported in the dataset, as a binary attribute, while birth years are separated into 6 equal-sized age range bins, and locations are translated from latitude/longitude pairs into Geohash strings with a precision of two characters, which results in 5 potential location classes. Details of the dataset and pre-processing steps can be found in Appendix A.

In our initial baseline experiment, we train a feature extraction module consisting of a pre-trained BERT model \citep{Devlin2019bert} along with two dense layers in order to extract useful features from the input text $x$. We obtain the final hidden state of the pre-trained model for each token in the input, then take a mean average over the sequence to produce an embedding for the full text, such that:
\begin{equation}
    x_e = f(x)
\end{equation}

Sentiment analysis is then carried out by a classifier  which learns to predict the review rating label $y$ given the embedding vector. Classifier setup and hyper-parameter details are listed in Appendix B.  

We simulate a task that an attacker may wish to perform on the input text by training a secondary classifier along with the target task that attempts to predict the value of private information variables $z$. Following \citet{Coavoux2018inference}, we target several features of the respondent as extracted from the dataset, namely gender, location, and birth year. These features, while in reality not being private by virtue of being public information provided by users, represent good proxies for sensitive attributes that users may not wish to be inferred from similar public datasets. In this sense, they provide a useful benchmark of the potential privacy risk, while allowing us to avoid unethical inferences concerning private attributes not shared by the user. 

\subsection{Adversarial Training}

In order to promote invariance in the text representation with respect to our private variables, we adopt the approach pioneered by \citet{Ganin2017domain}. Initially designed to promote domain-independent learning, this system involves training a secondary objective to predict features we do not wish to be distinguishable via gradient descent, then passing the loss through a gradient reversal layer into a target task objective, represented in our experiments by the feature extractor.

For a single instance of our data $(x_e, y, z)$ the adversarial classifier optimizes:
\begin{equation}
    \mathcal{L}_a(x_e,y,z;\theta_a) = -log P(z|x_e;\theta_a)
\end{equation}

Hence, the combination of both target and attacker classifiers lead to the following objective function, where $\theta_r$, $\theta_p$, $\theta_a$ represent the parameters of the feature extractor, classifier and adversarial classifier respectively:
\begin{equation}
\begin{aligned}
\mathcal{L}(x_e, y, z; \theta_r, \theta_p,\theta_a) = & -log P(y|x_e;\theta_r, \theta_p)\\
& -\lambda log P(\neg z|x_e;\theta_a)
\end{aligned}
\end{equation}

\noindent where $\neg$  indicates that the log likelihood of the private label $z$ is inverted, and $\lambda$ is the regularization parameter scaling the gradient from our adversarial classifier.  

\subsection{Embedding Perturbation}

Since it is also desirable to provide a measure of general privacy alongside the specific attacker task we simulate in our adversarial training, we adopt the local DP method of \citet{Lyu2020perturb} to perturb the feature representations we produce. Converting the generated embedding into a DP-compliant representation requires us to inject calibrated Laplace noise into the hidden state vector obtained from the pre-trained language model as follows:
\begin{equation}
\tilde{x}_e = x_e + n
\end{equation}

\noindent where $n$ is a vector of equal length to $x_e$ containing i.i.d. random variables sampled from the Laplace distribution centred around 0 with a scale defined by $\frac{\Delta f}{\epsilon}$, where $\epsilon$ is the privacy budget parameter and $\Delta f$ is the sensitivity of our function. 
 
Since determining the sensitivity of an unbounded embedding function is practically infeasible, we constrain the range of our representation to [0,1], as recommended by \citet{Shokri2015dp}. In this way, the L1 norm and the sensitivity of our function summed across $n$ dimensions of $x_e$ are the same, i.e. \( \Delta f = 1 \).

\begin{algorithm}[ht]
\SetAlgoLined
\SetKwInOut{KwIn}{Input}
\KwIn{Input data $x$, Label $y$, Private information label $z$ }
Extract features from input sequence: $x_e = f(x)$\;
Normalise representation: $x_e \leftarrow x_e - \min{x_e} / (\max{x_e} - \min{x_e}$)\;
Apply perturbation: $\tilde{x}_e = x_e + Lap(\frac{\Delta f}{\epsilon})$\;
Train classifiers: $\mathcal{L}(\tilde{x}_e, y, z; \theta_r, \theta_p) = -log P(y|\tilde{x}_e;\theta_r, \theta_p) - \lambda log P(\neg z|\tilde{x}_e;\theta_a)$
\caption{Context-Aware Private Embeddings (CAPE)}
\label{alg:CAPE}
\end{algorithm}

\subsection{Context-Aware Private Embeddings (CAPE)}

To preserve the general privacy benefits of DP-compliant embeddings with invariance to the specific private variable identified for adversarial training, we combine both processes in a system we call Context-Aware Private Embeddings (CAPE). Algorithm \ref{alg:CAPE} presents the joint adversarial training scheme with perturbed embedding sequences derived from our feature extractor.

\section{Evaluation and Results}
\label{sec:results}

\subsection{Evaluation}

We evaluate performance on the target task (i.e. sentiment analysis) and on our simulated attacker task (i.e. classifying each private attribute) with the F1-score metric. We do not provide base accuracy since it may not fully represent performance in an imbalanced multi-class setting. It should be noted that lower results for the attacker classifier denote greater empirical evidence of privacy (\textit{i.e.}, the attacker cannot predict the target variable). All evaluations were performed randomly selecting 70\% of the data for training (the remaining 30\% for testing). We compute mean and standard deviation of the F1-score over 4 runs.

\subsection{Results}

Table \ref{tab:comparison} shows the results for each system, with $\epsilon$ and $\lambda$ parameters static at 0.1 and 1.0 respectively. These values are derived from a set of experiments with a range of privacy parameter values as detailed in Appendix C.

\begin{table}[ht]
    \centering
    \begin{tabularx}{\columnwidth}{lYYYY}
    \toprule
         & \multicolumn{2}{c}{Target} & \multicolumn{2}{c}{Attacker}\\
         \cmidrule(lr){2-3} \cmidrule(lr){4-5}
         Approach & F1 & SD & F1 & SD\\
    \midrule
    Location &\\
    \midrule
    Base & \textbf{0.7759} & \textbf{0.0059} & 0.7850 & 0.0014\\
    Adv. & 0.7617 & 0.0051 & 0.7883 & 0.0093\\
    DP & 0.6367 & 0.0001 & 0.7806 & 0.0041\\
    CAPE & 0.6394 & 0.0014 & \textbf{0.7558} & \textbf{0.0093}\\
    \midrule
    Gender &\\
    \midrule
         Base & \textbf{0.7721} & \textbf{0.0025} & 0.7623 & 0.0056\\
         Adv. & 0.7719 & 0.0093 & 0.7517 & 0.0066\\
         DP & 0.6367 & 0.0001 & 0.7548 & 0.0036\\
         CAPE & 0.6419 & 0.0050 & \textbf{0.7325} & \textbf{0.0080}\\
    \midrule
    Age Range &\\
    \midrule
    Base & 0.7726 & 0.0078 & 0.2100 & 0.0177 \\
    Adv. & \textbf{0.7787} & \textbf{0.0068} & 0.0979 & 0.0041\\
    DP & 0.6367 & 0.0001 & 0.0528 & 0.0029\\
    CAPE & 0.6344 & 0.0083 & \textbf{0.0523} & \textbf{0.0017}\\
    \bottomrule
    \end{tabularx}
    \caption{Results for the target task and the simulated attacker task. CAPE outperforms all other approaches in terms of privacy-preservation for all variables.}
    \label{tab:comparison}
\end{table}

\section{Discussion and Conclusion}
\label{sec:concl}
These results demonstrate the enhanced privacy afforded by the CAPE approach over either privacy approach applied in isolation. We provide evidence that adversarial training can produce superior outcomes to a DP-only approach, if we consider the private variable targeted in training. Adding DP noise clearly harms performance outcomes, indicating that we require further work to implement alternate processes for perturbing embeddings. Perturbed embeddings generated in Euclidean space perform more poorly as the privacy guarantee increases, so projecting embeddings into Hyperbolic space \cite{Dhingra2018} or implementing a search mechanism to select semantically-similar vectors that represent real words \cite{Feyisetan2020embeds} could produce better outcomes with lower privacy budgets.

Interestingly, we find that different private attributes are predictable by an attacker at different rates---while the attacker can predict the correct gender or location class effectively, results for age range are barely above random chance. It may well be the case in the UK that word choice varies more between areas and genders than age cohorts, for example, a reviewer who cites the product's "lush vanilla taste" may reside in the West of England, while calling a bad service "shite" may indicate they are Scottish. This is an interesting counter-finding to \citet{welch-etal-2020-compositional} which found better embedding performance with age- and gender-aware representations in a global population. Differing privacy requirements for separate attributes are a feature of multiple variations on differential privacy regimes \citep{kamalaruban_not_2020, Alaggan_Gambs_Kermarrec_2017, jorgensen_conservative_2015}.

We note that English exhibits fewer grammatical markers that indicate gender than some other languages \citep{boroditskySexSyntaxSemantics2000}, a peculiarity which may affect the utility of the model in significant ways. Future work will focus on additional European languages.

\pagebreak

\bibliographystyle{acl_natbib}
\bibliography{references, acl2021}

\end{document}